\documentclass[journal]{IEEEtran}
\usepackage{amsmath}
\usepackage{amssymb}
\usepackage{booktabs} 
\usepackage{xcolor}
\usepackage{algpseudocode}
\usepackage{algorithm,tabularx}
\usepackage{multirow}
\usepackage{helvet}  
\usepackage{courier}  
\usepackage{url}  
\usepackage{epstopdf,subfigure}
\usepackage{amscd,epsfig,amsfonts,rotating}
\usepackage{algpseudocode}
\usepackage{algorithm,tabularx}

\makeatletter
\newcommand{\multiline}[1]{%
  \begin{tabularx}{\dimexpr\linewidth-\ALG@thistlm}[t]{@{}X@{}}
    #1
  \end{tabularx}
}

\newcommand{\bs}{\boldsymbol}
\newcolumntype{C}[1]{>{\centering\arraybackslash}p{#1}}
\graphicspath{{fig/}}

\begin{document}
%
\title{Point2SpatialCapsule: Aggregating Features and Spatial Relationships of Local Regions on Point Clouds using Spatial-aware Capsules}
%
%
%

\author{Xin Wen,
        Zhizhong Han,
        Xinhai Liu,
        Yu-Shen Liu,~\IEEEmembership{Member,~IEEE}
\thanks{X. Wen and X. Liu are with the School of Software, Tsinghua University, Beijing 100084, China (e-mail: x-wen16, lxh17@mails.tsinghua.edu.cn).}
\thanks{Z. Han is with the School of Software, Tsinghua University, Beijing 100084, China, and also with the Department of Computer Science, University of Maryland at College Park, College Park, MD 20737 USA (e-mail: h312h@mail.nwpu.edu.cn).}
\thanks{Y.-S. Liu is with the School of Software, Tsinghua University, Beijing 100084, China, and also with the Beijing National Research Center for Information Science and Technology (BNRist), China (e-mail: liuyushen@tsinghua.edu.cn). (Corresponding author: Yu-Shen Liu.)}
\thanks{This work was supported by National Key R\&D Program of China (2018YFB0505400)}}

\markboth{Journal of \LaTeX\ Class Files,~Vol.~14, No.~8, August~2019}%
{Shell \MakeLowercase{\textit{et al.}}: Bare Demo of IEEEtran.cls for IEEE Journals}
%

\maketitle

\begin{abstract}
Learning discriminative shape representation directly on point clouds is still challenging in 3D shape analysis and understanding. Recent studies usually involve three steps: first splitting a point cloud into some local regions, then extracting the corresponding feature of each local region, and finally aggregating all individual local region features into a global feature as shape representation using simple max pooling. However, such pooling-based feature aggregation methods do not adequately take the spatial relationships (e.g. the relative locations to other regions) between local regions into account, which greatly limits the ability to learn discriminative shape representation. To address this issue, we propose a novel deep learning network, named Point2SpatialCapsule, for aggregating features and spatial relationships of local regions on point clouds, which aims to learn more discriminative shape representation. Compared with the traditional max-pooling based feature aggregation networks, Point2SpatialCapsule can explicitly learn not only geometric features of local regions but also the spatial relationships among them. Point2SpatialCapsule consists of two main modules. To resolve the disorder problem of local regions, the first module, named \emph{geometric feature aggregation}, is designed to aggregate the local region features into the learnable cluster centers, which explicitly encodes the spatial locations from the original 3D space. The second module, named \emph{spatial relationship aggregation}, is proposed for further aggregating the clustered features and the spatial relationships among them in the feature space using the spatial-aware capsules developed in this paper. Compared to the previous capsule network based methods, the feature routing on the spatial-aware capsules can learn more discriminative spatial relationships among local regions for point clouds, which establishes a direct mapping between log priors and the spatial locations through feature clusters. Experimental results demonstrate that Point2SpatialCapsule outperforms the state-of-the-art methods in the 3D shape classification, retrieval and segmentation tasks under the well-known ModelNet and ShapeNet datasets.
%
\end{abstract}

\begin{IEEEkeywords}
point cloud, shape representation, feature aggregation, spatial relationships, capsule network.
\end{IEEEkeywords}

\IEEEpeerreviewmaketitle

\section{Introduction}

3D shape representation learning plays a central role in shape analysis and understanding, which has a wide range of applications such as shape classification \cite{qi2017pointnet,han20193dviewgraph,komarichev2019cnn}, retrieval \cite{han2019view,han2019parts,han2019seqviews2seqlabels}, semantic segmentation \cite{liu2019sequence,lei2019octree} and instance segmentation \cite{wang2019associatively,hou20193d}.
Among the multiple representation forms of 3D shapes, 3D point clouds, benefited from its easy access, have become one of the most popular 3D shape forms in recent years. Specifically, the point clouds consist of a set of unordered points, each of which is composed of 3D coordinates, possibly with some additional attributes such as normal, color and material.

%
\begin{figure}[!tp]
  \centering
  \includegraphics[width=\linewidth]{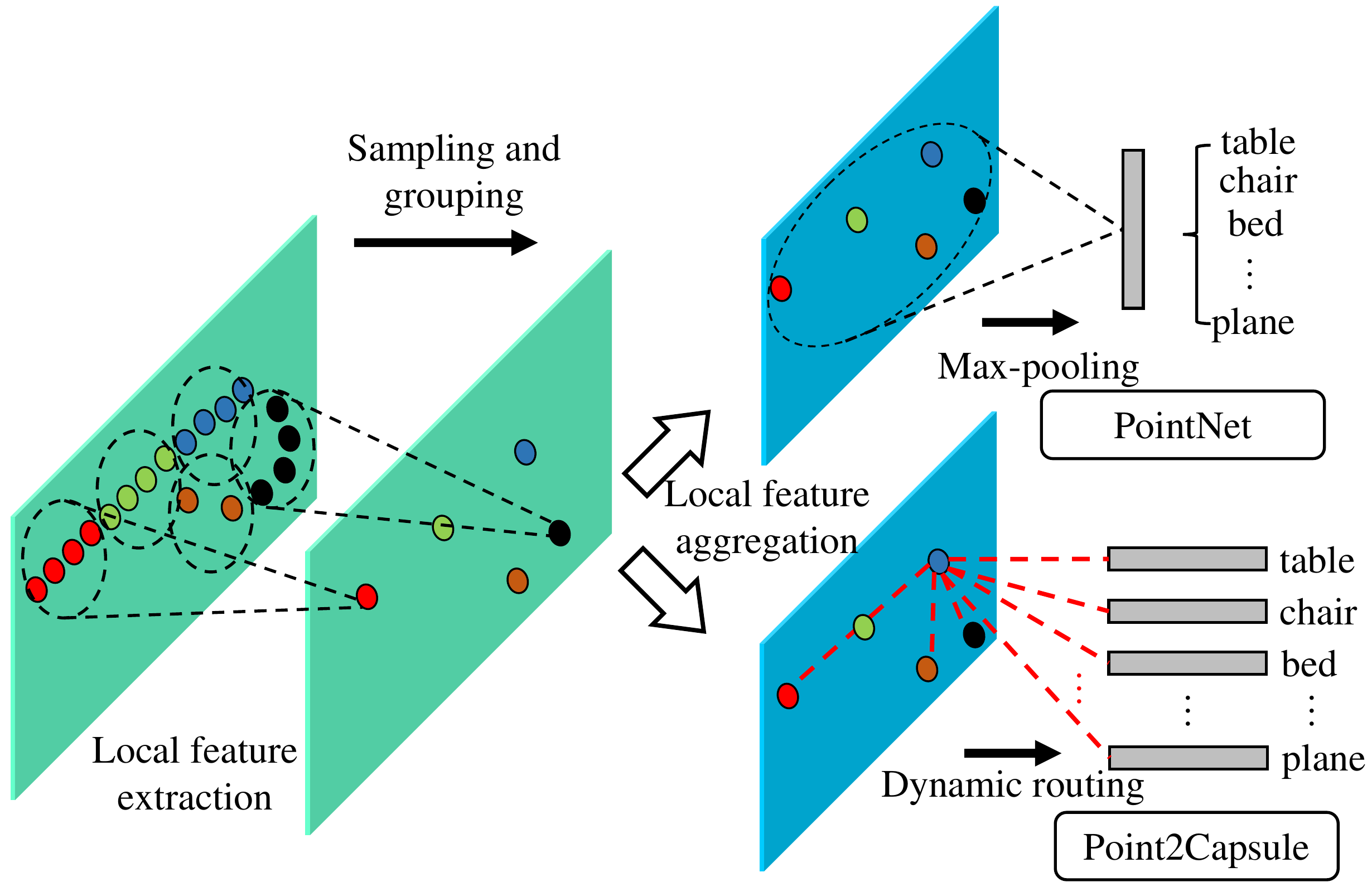}
 \caption{The illustration of comparison between the max-pooling based PointNet and the dynamic routing based Point2SpatialCapsule in terms of local region feature aggregation. Given an example 2D point cloud like a shape of ``P'', the point cloud is split into some local regions (left), from which the corresponding local region features are extracted with sampling and grouping (middle). The comparison of two methods is shown at the right side of this figure. Here, denoted by black dotted line, the max-pooling only keeps the most significant geometric characteristics in local regions (top right), while this causes the spatial relationships between local regions are filtered out. In contrast, the dynamic routing based Point2SpatialCapsule can handle both the geometric characteristics and the spatial relationships of local regions (bottom right), denoted by the red dotted line.
  }
  \label{fig:comparison}
\end{figure}

However, learning discriminative shape representation directly on point clouds is still challenging in 3D shape analysis and understanding. Recent studies for learning point cloud representations usually involve the following three steps. Each input point cloud is first split into some local regions. Then, the corresponding features of local regions are extracted using shared Multi-Layer Perceptron (MLP) \cite{qi2017pointnet} or kd-trees \cite{klokov2017escape}. Finally, the extracted local region features are aggregated into a global feature vector as the shape representation \cite{qi2017pointnet++,li2018pointcnn,liu2019sequence}.
Most of the previous methods mainly focus on how to enhance the process of local region feature extraction, while often employ a simple pooling-based layer \cite{qi2017pointnet,qi2017pointnet++,li2018pointcnn} to aggregate these extracted features. However, such pooling-based feature aggregation methods do not take adequately the spatial relationships among local regions into account.
So far, how to aggregate those learned local region features and their spatial relationships still remains the challenges in existing methods of point cloud representation learning.
In this paper, we first argue the importance of learning spatial relationships for aggregating local region features with respect to the following two reasons. (1) For point clouds with similar local regions, the differences in the spatial arrangements of these local regions are important for learning the discriminative features. (2) Considering the permutation invariant nature of point clouds, it is important to learn the intrinsic spatial relationship between each part and the whole, in order to constitute the permutation invariant knowledge for point cloud recognition.

%
\begin{figure}[!tp]
  \centering
  \includegraphics[width=\linewidth]{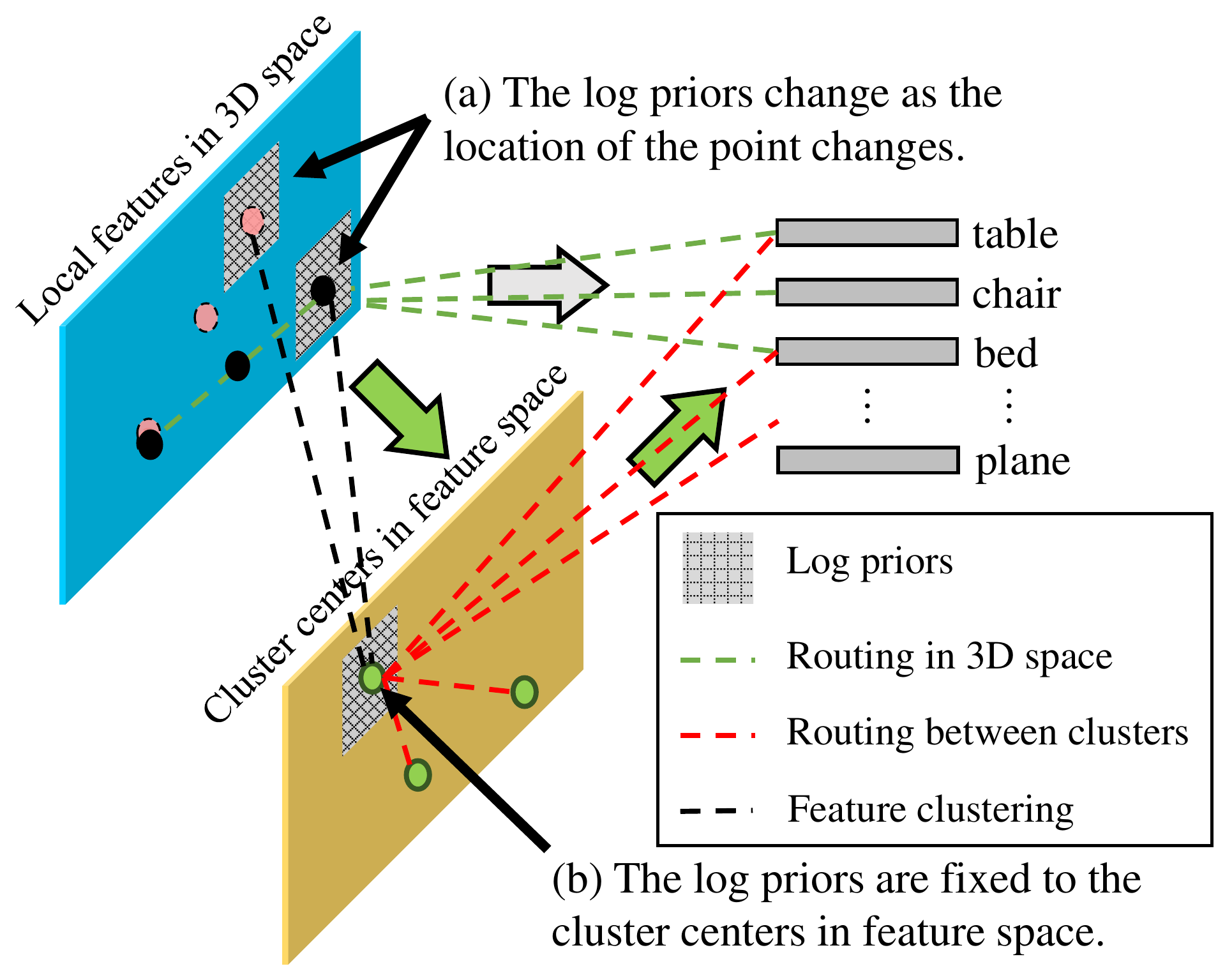}
 \caption{The illustration of directly applying capsule network to the point cloud (a) and the clustering based Point2SpatialCapsule (b). The shifting and rotation of point cloud change the locations of local regions in 3D space and also change their corresponding log priors. Therefore, routing in 3D space (green dotted line) will cause the shifting of log priors, making the routing algorithm fail to learn the spatial relationships between local regions.
  In contrast, the geometric feature aggregation aggregates the input local region features into relatively invariant cluster centers. Therefore, routing between clusters (red dotted line) can efficiently learn the log priors for aggregating spatial relationships between local regions.}
  \label{fig:why_netvlad}
\end{figure}

The common strategy for feature aggregation in previous methods \cite{qi2017pointnet++,qi2017pointnet,li2018so,shen2018mining} is to extract the most significant characteristics (such as the engine of an airplane) in local regions of the point cloud step by step, through a deep neural network with the pooling structure (such as max-pooling).
However, the problem is that the pooling-based methods will filter out the spatial relationships of different areas on the feature map \cite{sabour2017dynamic}. Thus, it only considers the existences of characteristics in local regions, while the spatial arrangements between these regions will not be preserved.
As a result, most of the existing methods usually fail to learn the spatial relationships among the local regions, which further limits the ability of the network for learning discriminative 3D shape representation.

To address the aforementioned problem, we propose a novel deep learning network, named Point2SpatialCapsule, for aggregating geometric features and spatial relationships of local regions on point clouds, which aims to learn more discriminative shape representation. Inspired by the recently developed capsule network \cite{sabour2017dynamic}, Point2SpatialCapsule employs the dynamic routing to aggregate the local region features and their spatial relationships. Fig. \ref{fig:comparison} illustrates the comparison between our dynamic routing based Point2SpatialCapsule and the previous feature aggregation methods like max-pooling used in PointNet \cite{qi2017pointnet}. Note that the max-pooling feature aggregation in PointNet \cite{qi2017pointnet} only considers the existences of characteristics in local regions, but in contrast our Point2SpatialCapsule can explicitly handle the spatial relationships between local region features through dynamic routing of capsule network. This advantage encourages us to consider adopting capsule network for 3D point clouds representation learning.


However, the problem is that, the original implementation of capsule network is designed for 2D image recognition, where the log priors in capsule network are bounded to the fixed locations on the 2D feature maps \cite{sabour2017dynamic}. In contrast, for 3D point clouds, the locations of random sampled input points are disordered and their absolute position coordinates may not always keep consistent. As a result, it is difficult to find a direct mapping that can generate the features encoded with fixed spatial locations.
What's worse, the previous capsule based methods failed to address such problem, most of which directly generate the capsules from a single global feature vectors. Such practice leads to the loss of spatial relationships between local regions. As a result, the log priors in routing algorithm between capsules can not learn the spatial relationships of local regions, which greatly limits the representation ability of capsules.
In this paper, we argue the importance for encoding the fixed spatial locations into capsules, which aims to efficiently utilize the representation ability of log priors for learning the spatial relationships between local regions on point clouds.

In order to solve the above limitations, two novel modules are specially designed in Point2SpatialCapsule to achieve local region feature aggregation as follows. (1) The first module, named \emph{geometric feature aggregation}, aims to aggregate the extracted local region features in the feature space. Here, the term ``\emph{geometric}'' indicates that this module aggregates the geometric information, like the coordinates of central points and the shapes of local regions represented by the feature vectors, into the centers of local feature clusters, which aims to resolve the disorder problem of local regions. (2) The second module, named \emph{spatial relationship aggregation}, is to apply routing algorithm on the learned feature clusters.
The term ``\emph{spatial-aware}'' indicates that the capsules are encoded with the spatial locations, which is to guarantee the direct mapping between the log priors and the fixed locations in the 3D space. Therefore, we call them the \emph{spatial-aware capsules}, which allows the network to efficiently learn the spatial relationships between local regions.
Fig. \ref{fig:why_netvlad} shows the visualized demonstration of the advantage of spatial relationship aggregation. Because of the shifting and rotation of point clouds, the changing locations of local region features in 3D space also change the log priors.
To resolve this issue, the geometric feature aggregation clusters the input local region features into the learnable cluster centers, which are irrelevant to the input points and relatively invariant in the feature space. Therefore, the routing algorithm can efficiently learn the log priors for aggregating the spatial relationships between local regions.
Our main contributions are summarized as follows.
\begin{itemize}
  \item We propose a novel deep network, i.e. Point2SpatialCapsule, for learning more discriminative shape representations of point clouds. Compared with the traditional pooling-based methods, Point2SpatialCapsule can explicitly learn not only geometric features of local regions but also the spatial relationships among them.
  \item We propose the geometric feature aggregation to resolve the disorder problem of local regions, where the local region features are aggregated into the learnable cluster centers, which are explicitly encoded with the spatial locations from the original 3D space.
  \item We propose the spatial relationship aggregation to further utilize the spatial locations encoded in the feature clusters. Compared to the previous capsule network based methods, the spatial relationship aggregation can learn more discriminative spatial relationships between local regions by establishing a direct mapping between log priors and the spatial locations through feature clusters.
\end{itemize}

The remainder of this paper is organized as follows. First, the related work is introduced in Sec.\ref{sec:related_work}. Then we detail the proposed Point2SpatialCapsule in Sec.\ref{sec:model_description}. The experiments and the ablation studies are given in Sec.\ref{sec:experiments}. Finally, we conclude this paper in Sec.\ref{sec:conclusions}.

\section{Related Work}
\label{sec:related_work}

In this section, we mainly review the methods related to 3D shape representation learning based on deep learning networks. The existing methods can be roughly divided into four categories according to various 3D shape forms that are learned from, including voxels, point clouds, views and meshes.

\subsection{Point Cloud Based Methods}

Recent studies of point cloud representation learning mainly focus on the local feature extraction and integration. PointNet \cite{qi2017pointnet} is the pioneering work of introducing deep learning into point cloud representation learning,  which independently learns the features of each point and aggregates the learned features into a global feature with the max-pooling layer. After that, plenty of the follow-up studies \cite{li2018pointcnn,liu2019sequence,xu2018spidercnn,wang2018dynamic} focus on how to better integrate the contextual information of local regions on point clouds.
For example, PointNet++ \cite{qi2017pointnet++} designed the hierarchical feature learning architecture based on PointNet to encode multi-scale local areas. Following the convolutional structure of PointNet++, successors such as PointCNN \cite{li2018pointcnn} and SpiderCNN \cite{xu2018spidercnn} investigated some improved convolution operations which aggregate the neighbors of a given point by edge attributes in the local region graph.
Different from the idea of using convolution structure, Point2Sequence \cite{liu2019sequence} introduced the sequential model (i.e. RNN) to capture the fine-grained contextual information of features in local regions. Specifically, Point2Sequence arranges the features into a sequence according to the size of the region scale, and then uses a RNN to capture the contextual information within the local regions.
However, the problem is that most of the above methods fail to consider the spatial relationships among different local regions when aggregating the extracted local region features, where the usual practice for these methods is to use the pooling layer to learn the global feature from the local ones.

More recent studies focus on how to improve the local region feature extraction \cite{komarichev2019cnn,liu2019relation,zhao2019pointweb}. These methods have shown impressive potentials in the semantic segmentation task on point cloud. For examples, A-CNN \cite{komarichev2019cnn} was proposed to annularly arrange the neighbor points and apply the convolution network on these arranged points to learn the local region features. RS-CNN \cite{liu2019relation} designed a shape-aware convolution to learn the local region features from the relation between points.

The proposed Point2SpatialCapsule mainly focus on how to aggregate the feature and relationships of local regions after extracting local features. The usual practice for previous methods is to apply the strategy of bottom-to-top point cloud feature aggregation \cite{klokov2017escape,wang2017cnn,li2018so,lei2019octree,riegler2017octnet}. For example, Kd-Net \cite{klokov2017escape} performs multiplicative transformations according to the subdivisions of point clouds based on the kd-trees.
SO-Net \cite{li2018so} employs a SOM to build the spatial distribution of the input point cloud, which allows hierarchical feature extraction on both individual points and SOM nodes.
However, most of the above methods use max-pooling as a feature aggregation method, which inevitably filters out the spatial relationships among local regions.
On the other hand, PVNet \cite{you2018pvnet} is also a notable method that considers the local feature aggregation, which focuses on mining the difference in importance between the local features.
It employs high-level global features from the multi-view data of input 3D shapes to mine the relative correlations between different local features from the point cloud data. Same as the above-mentioned methods, PVNet only learns the different contributions among local regions, while the spatial relationships among these regions are not considered.

\subsection{View-based Methods}

The dominant performance of multi-view based methods on the task of 3D shape retrieval comes from the research progress of measuring the similarities between 2D image features \cite{han20193d2seqviews,bai2017gift,han2019multi,han20182seq2seq}. As one of the pioneering work, GIFT \cite{bai2017gift} adopted the Hausdorff distance to measure the similarity between the view sets of two 3D shapes. Another notable research direction is to focus on PANORAMA views of 3D shapes, where a PANORAMA view can be regarded as the seamless aggregation of multiple views captured on a circle. For examples, DeepPano \cite{shi2015deeppano} introduced a row-wise max-pooling to relief the effect of rotation about the up-oriented direction, and Sfikas et al. \cite{sfikas2017panorama} introduced CNN for learning the global features from the PANORAMA views in a consisitent order. To explore the potential of attention mechanism, the methods like 3DViewGraph \cite{han20193dviewgraph} have been proposed to integrate the spatial pattern correlations of unordered views with attention weights, and Part4Features \cite{han2019parts} developed a novel multi-attention mechanism for aggregating the learned local parts.

More recently, SeqViews2SeqLabels \cite{han2019seqviews2seqlabels} was proposed to learn 3D features via aggregating sequential views by RNN, which aims to eliminate the effect of rotation of 3D shapes. Compared with the previous pooling based methods, the RNN-based SeqViews2SeqLabels suffers less from the content and the spatial location loss. Similarly, as an unsupervised approaches, VIP-GAN \cite{han2019view} trains an RNN-based neural network architecture to solve multiple view inter-prediction tasks for each shape.

\subsection{Voxel-based Methods}

Voxel-based methods often rasterize a 3D shape as a  function or distribution sampled on voxels \cite{Liu11pami,Liu12cad}.
For supervised learning the representation of 3D voxels, 3DShapeNets \cite{wu20153d} adopted the convolutional restricted Boltzmann machine to learn the representation of 3D voxels. O-CNN \cite{wang2017cnn} learns the representation of 3D voxel based on a novel octree structure. And Han et al. \cite{HanCyber17a} proposed a novel permutation voxelization strategy to learn high-level and hierarchical 3-D local features from raw 3-D voxels. For unsupervised learning, methods like VConv-DAE \cite{sharma2016vconv} use the fully convolutional autoencoder for unsupervised learning the voxel representation by reconstruction.
However, the problem is, considering the induced complexity and limitations of directly exploiting the sparsity of voxel grids, it is difficult to introduce the large scale or flexible deep networks for representation learning.
Therefore, more recent methods such as OctNet \cite{riegler2017octnet} and kd-net \cite{klokov2017escape} consider to utilize the scalable indexing structures for solving this problem, where deep neural networks can be further adopted for achieving more impressive results.

\subsection{Mesh-based Methods}

As for mesh-based methods, to explore the effectiveness of the heat diffusion based descriptor, Xie et al. \cite{xie2015deepshape} proposed a shape feature learning scheme based on auto-encoders, where the model can extract the features that are insensitive to the deformations. By fully utilizing the spectral domain, Xie et al. \cite{xie2016learned} further proposed to learn a novel binary spectral shape descriptor with the deep neural network for 3D shape correspondence. Recently,
BoSCC \cite{Liu17TIP-BoSCC} was introduced for a spatially
enhanced 3D shape representation based on bag of spatial context
correlations. And more recently, Deep Spatiality \cite{han2018deep} was also proposed to simultaneously learn 3D global and local features with novel coupled softmax.

\subsection{Capsule Networks}

The ability of capsule network \cite{sabour2017dynamic} for capturing spatial relationships comes from the dynamic routing algorithm and the log priors, which are bound to the absolute location on the input feature maps.
Specifically, the capsule network learns the log priors by considering the relationships between the absolute locations on the feature map and the high-level capsules. Then, through the dynamic routing algorithm, which is based on the learned log priors, the high-level capsules can integrate the low-level features and their spatial relationships among different locations on the feature maps. This advantage promotes us to consider applying the capsule network to 3D point cloud representation learning.

So far, the capsule network has shown the great potentials in many research areas, such as image processing \cite{jaiswal2018capsulegan,lalonde2018capsules} and natural language processing (NLP) \cite{yang2018investigating,xiao2018mcapsnet,ning2018capsule}.
However, as for the application of capsule network in 3D shape representation learning, there are a few methods proposed in recent years. For example, 3D-CapsNet \cite{burak2018capsules} adopts the capsule network for 3D shape classification tasks based on volumetric data, and 3D-Point-Capsule \cite{zhao2018capsule} learns the point cloud representation and part segmentations in an unsupervised way. And for supervised learning, 3DCapsule \cite{che2019capsule} applies the capsule network as an extension of fully-connected layers for point cloud classification.

An important problem of the above methods is that they all build the capsule layers over the global feature (usually produced by the fully-connected layer or max-pooling) of point clouds, where the spatial relationships between local region features have been filtered out by the network. Therefore, the log priors in routing algorithm cannot learn the spatial distribution among the extracted local features, which limits the biggest advantage of capsule network for aggregation spatial relationships of local regions.

Therefore, to address this problem of previous methods, Point2SpatialCapsule aggregates the features into clusters in feature space, and applies the routing algorithm between these aggregated clusters. In the research of point cloud representation learning, methods like PointNetVLAD \cite{mik2019pointnetvlad} have adopted the similar clustering strategy, i.e. NetVLAD \cite{arandjelovic2016netvlad}, for feature aggregation. However, different from the previous methods that only cluster features for aggregating regions with similar geometric characteristics (e.g. shapes), our method takes one step further to not only considering geometric characteristics, but also explore the potentials for aggregating spatial relationships between these regions.
Specifically, Point2SpatialCapsule produces the clusters for both the features and their coordinates, in order to explicitly preserve the features and their spatial location.

\begin{figure*}[!tp]
  \centering
  \includegraphics[width=\textwidth]{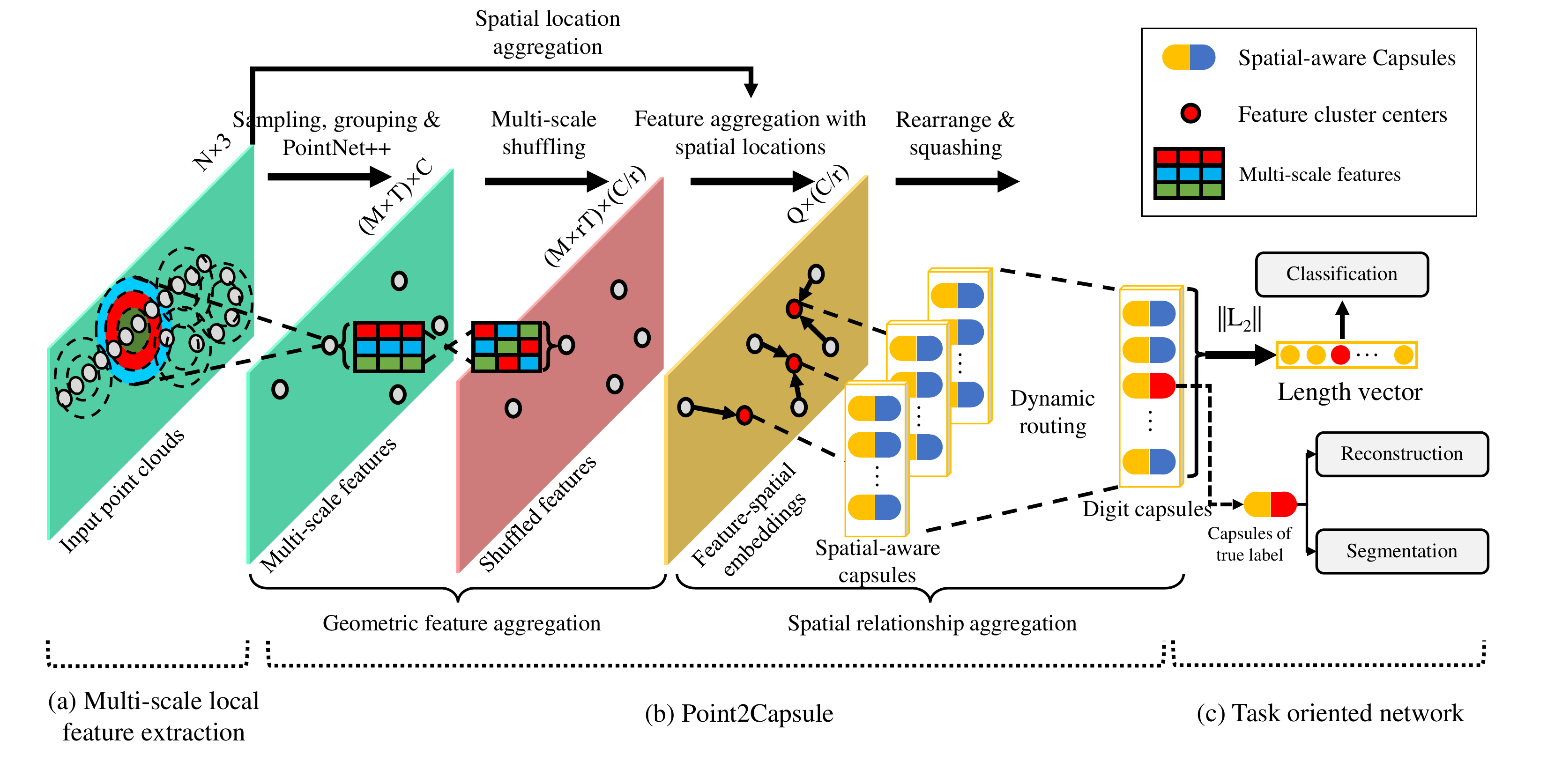}
  \caption{The architecture of our proposed Point2SpatialCapsule. For input point clouds, (a) the multi-scale local feature extraction first extracts features from multi-scale areas, (b) then the geometric feature aggregation encodes the extracted multi-scale local region features and their locations into the learnable clustering centers to produce the feature-spatial embeddings. The spatial relationship aggregation aggregates the feature-spatial embeddings by considering both the embeddings and their spatial relationships. (c) The task oriented network is adopted for performing on different tasks.}
  \label{fig:overview}
\end{figure*}

\section{Shape Representation Learning with Point2SpatialCapsule}
\label{sec:model_description}

An overview of shape representation learning network with Point2SpatialCapsule is shown in Fig. \ref{fig:overview}. The whole network consists of three main parts as follows. (1) The first part is the {multi-scale local feature extraction}, which is a PointNet++ based network for extracting the features from multi-scale local regions on point clouds (see Sec. \ref{sec:model_description:1}). (2) The second part is Point2SpatialCapsule, which is composed of two main modules for aggregating the learned features into the global shape representation. Here, the first module, i.e. \textit{geometric feature aggregation}, is to aggregate local region features into clusters (see Sec. \ref{sec:model_description:2}). The second module, i.e. \textit{spatial relationship aggregation}, is to aggregate the feature clusters and their spatial relationships into global feature representation (see Sec. \ref{sec:model_description:3}). In this section, we will also detail the training procedure of Point2SpatialCapsule (see Sec. \ref{sec:model_description:4}). (3) The third part is the task oriented network used for various tasks such as shape segmentation (see Sec. \ref{sec:model_description:5}).


\subsection{Multi-scale Local Feature Extraction}
\label{sec:model_description:1}

The first part of our network is the multi-scale local feature extraction, as shown in Fig. \ref{fig:overview}(a). Given a set of input points $\mathbf{X}=\{\bs{x}_1, \bs{x}_2,..., \bs{x}_{n} \}$, by following the practice of PointNet++ \cite{qi2017pointnet++} and ShapeContextNet \cite{xie2018attentional}, we iteratively produce a subsampling $\{\bs{x}_{k_1}, \bs{x}_{k_2},..., \bs{x}_{k_M} \}$ with $M$ points as the centroids of the local regions using farthest point sampling (FPS), such that the newly added point $\bs{x}_{k_j}$ is the farthest point (in metric distance) from the rest sampled points $\{\bs{x}_{k_1}, \bs{x}_{k_2},..., \bs{x}_{k_{j-1}} \}$. Then, for each sampled point, the $K$ nearest neighbor (kNN) searching is employed to find $\{K_i|i=1,...,T\}$ neighbors for this point, under $T$ different scale areas. Followed by a grouping layer, the sampled point and its neighbors are grouped as a $K_i \times 3$ tensor for scale $K_i$. After that, a simple but effective MLP layer is employed to extract the features of all neighbor points, producing a tensor with shape $K_i \times C$. Finally, a max-pooling layer is applied to integrate the point features in each scale to produce the scale feature of dimension $C$ for scale $K_i$. For $M$ points in total and $T$ scales for each point, the multi-scale local feature extraction layer produces $M \times T$ multi-scale features, forming a tensor of shape $M \times T \times C$ as its output.

In the implementation, we apply two layers of multi-scale local feature extraction for hierarchically extracting features from point clouds.


\subsection{Point2SpatialCapsule: Geometric Feature Aggregation}
\label{sec:model_description:2}

In this subsection, we detail the first module of Point2SpatialCapsule, which aims to aggregate the extracted features into clusters and encodes these features with spatial locations.

As shown in Fig. \ref{fig:overview}(b), before clustering features, the module of geometric feature aggregation first applies the multi-scale shuffling to enhance the diversity of features. Then the features are aggregated into clusters and encoded with the spatial locations (e.g. the absolute locations in the 3D space) from the original 3D space.

\begin{figure}[!t]
  \centering
  \includegraphics[scale=0.8]{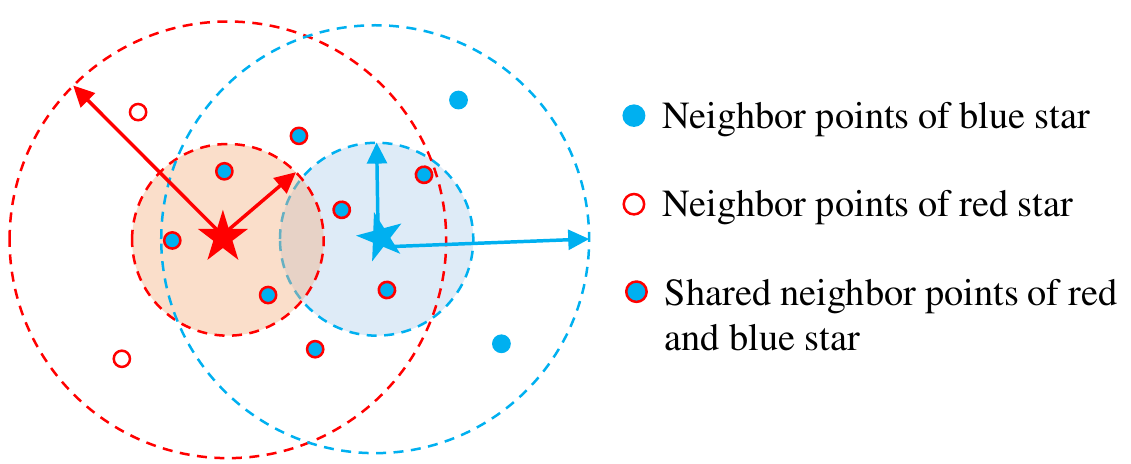}
  \caption{Illustration of feature similarity between adjacent points. Features of small scales are not very similar because of the small overlap in sampled points. On the contrary, the features of large scales are more similar because of a larger overlap.}
  \label{fig:scale-cmp}
\end{figure}

\subsubsection{Multi-scale Shuffling}

Different from the previous methods that apply the pooling-based strategy for integrating the features extracted from multi-scale regions, we propose the multi-scale shuffling layer to build the shuffled features.
The reason for adding this layer is demonstrated in Fig. \ref{fig:scale-cmp}, as explained below. When searching the neighbor points in a large scale, the searching areas of two adjacent centroids will overlap with each other and output the same neighbor points. As a result, the adjacent points will tend to have the similar features for large scale, which can reduce the diversity of features and introduce an initial clustering center for the subsequent clustering layer.
On the other hand, the features of small scales between two centroids are dissimilar because of small overlaps. Therefore, the multi-scale shuffling is introduced to smooth the perceived range of features between different scales and enhance the feature diversity, by mixing the dissimilar features of small scales with the similar features of large scales. As a result, the multi-scale shuffling can promote the network to consider all input features equally and alleviate the problem of similar features.

\begin{figure}[!t]
  \centering
  \includegraphics[scale=0.8]{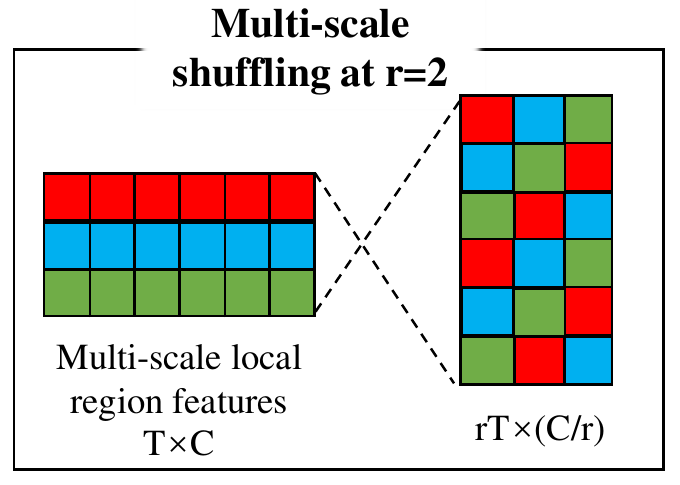}
  \caption{Illustration of the multi-scale shuffling, which is the solution to the problem of feature similarity.}
  \label{fig:multiscale_shuffle}
\end{figure}

The effect of multi-scale shuffling is shown in Fig. \ref{fig:multiscale_shuffle}. Specifically, given a point with $T$ scale features of $C$ dimension, which forms a tensor with the shape $T\times C$, the multi-scale shuffling periodically rearranges the elements in the $T\times C$ tensor into a tensor of shape $rT\times (C/r)$, where $r$ is an integer. Thus, for $M$ points in total, the multi-scale shuffling will produce $M\times rT$ shuffled features of dimension $C/r$, resulting in a tensor of size $(M\times rT)\times (C/r)$.

The multi-scale shuffling is inspired by the subpixel convolution \cite{shi2016real} for image upsampling, where the number of area scales $T$ can be considered as the size of image, and the feature dimension $C$ can be regarded as the channels of feature maps.
However, different from subpixel convolution which is designed for speeding up the calculations and reducing the amount of parameters in the network, the multi-scale shuffling used in our method aims to enhance the diversity of scale features.
We will quantitatively explore the importance of the multi-scale shuffling in ablation studies in Sec. \ref{subsec:ablation}.

\subsubsection{Feature Aggregation with Spatial Encodings}

\begin{figure*}[!tp]
  \centering
  \includegraphics[width=\textwidth]{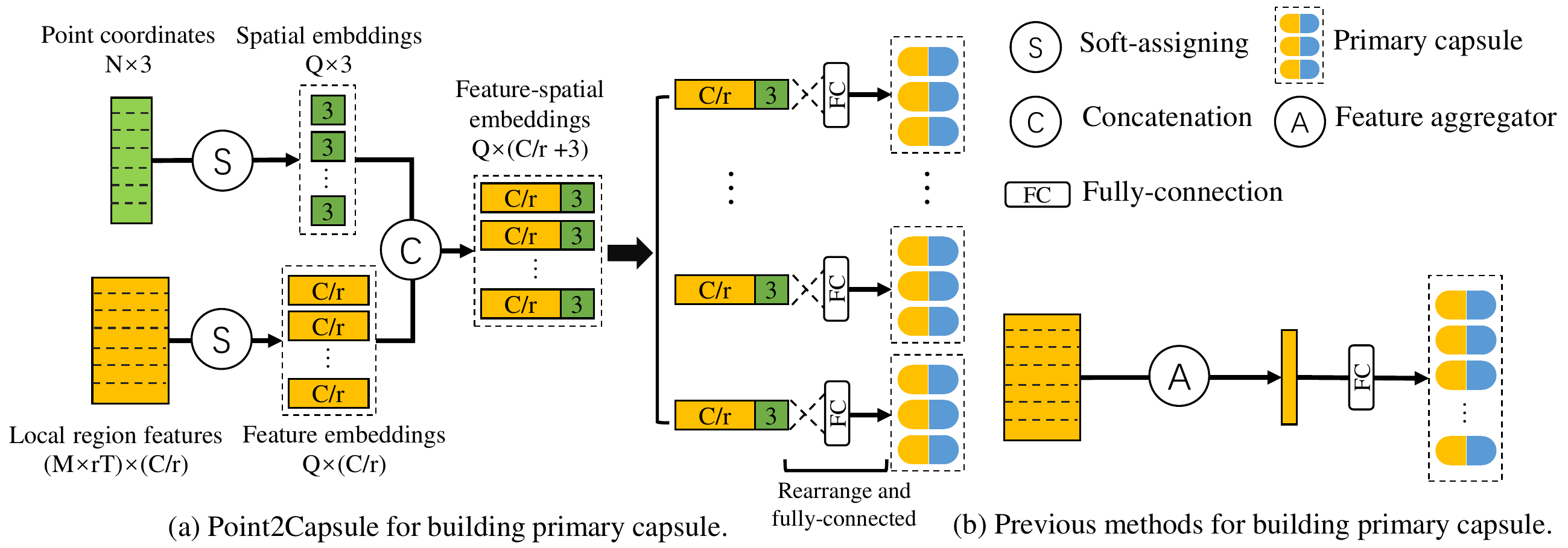}
  \caption{Comparison of the strategies for applying dynamic routing in local region features between (a) Point2SpatialCapsule and (b) the previous methods \cite{zhao2018capsule,che2019capsule}. The previous methods directly use a feature aggregator (e.g. max-pooling) to aggregates all local region features, and generate the capsules based on the aggregated global feature. In contrast, Point2SpatialCapsule proposes to cluster the local region features and the point coordinates, and then combines them as spatial-aware cluster centers. The spatial-aware capsules of Point2SpatialCapsule are independently generated according to each cluster centers.}
  \label{fig:comparison_vlad}
\end{figure*}

The purpose of this layer is to aggregate the shuffled features into the learnable feature cluster centers, which can be regarded as the latent embeddings describing the semantic patterns of the local regions features. To achieve this purpose, we propose to cluster the features in the feature space and their coordinates in the original 3D space. After that, the cluster centers in both the feature space and the 3D space are fused to produce \emph{the feature-spatial embeddings}, as illustrated in Fig. \ref{fig:comparison_vlad}(a).

Although the traditional clustering methods like k-means can be adopted to produce the feature cluster centers, their computational cost may be very high because of the huge number of features to be clustered. Therefore, inspired by the recent development of NetVLAD \cite{arandjelovic2016netvlad}, we adopt the soft-assignment for learning the clustering centers for the input shuffled local features. Specifically, the network learns $Q$ cluster centers for input features, denoted as $\{\bs{q}_1, \bs{q}_2,...,\bs{q}_Q | \bs{q}_k \in \mathbb{R}^{C/r}\}$, as colored by yellow in Fig. \ref{fig:comparison_vlad}(a). For each cluster center $\bs{q}_k$, the layer produces a \emph{feature embedding} $C(\bs{q}_k)\in \mathbb{R}^{C/r}$, which is an aggregated representation over the whole input shuffled features $\{\bs{\hat{p}}_i\}$, denoted by
\begin{equation}
  C(\bs{q}_k) = \sum_{i=1}^{n}\frac{e^{\bs{w}_k^T \bs{\hat{p}}_i}+b_k}{\sum_{k'}e^{\bs{w}_{k'}^T \bs{\hat{p}}_i}+b_{k'}}(\bs{\hat{p}}_i - \bs{q}_k),
\end{equation}
where $\{\bs{w}_k\}$ and $\{b_k\}$ are the weights and biases, respectively, that determine the contribution of each local feature to the cluster center $\bs{q}_k$. During training, all the weights, biases and the cluster centers are updated through back-propagation algorithm.

To explicitly encode the spatial locations of local features into their cluster centers, we first cluster the coordinates $\{\bs{x}_i\}$ of input points into the coordinates cluster centers $\{\bs{y}_1, \bs{y}_2,...,\bs{y}_Q | \bs{y}_k \in \mathbb{R}^{C/r}\}$, which is the same process as described above and colored by green in Fig. \ref{fig:comparison_vlad}(a). The \emph{spatial embeddings} $C(\bs{y}_k)\in \mathbb{R}^{C/r}$ for coordinates is given as
\begin{equation}
  C(\bs{y}_k) = \sum_{i=1}^{n}\frac{e^{\bs{w}_k^T \bs{x}_i}+{b'}_k}{\sum_{k'}e^{\bs{w}_{k'}^T \bs{x}_i}+{b'}_{k'}}(\bs{x}_i - \bs{y}_k).
\end{equation}
Then, the produced local feature embedding and its corresponding spatial embedding are concatenated to form an explicit \emph{feature-spatial embedding} $C(\bs{s}_k) = [C(\bs{y}_k):C(\bs{x}_k)]$.
%

\subsection{Point2SpatialCapsule: Spatial Relationship Aggregation}
\label{sec:model_description:3}

In Fig. \ref{fig:comparison_vlad}(b), we show the overall architecture of previous methods \cite{zhao2018capsule,che2019capsule} for building the capsules, and compare it with our proposed Point2SpatialCapsule shown in Fig. \ref{fig:comparison_vlad}(a). The main difference is that Point2SpatialCapsule builds the spatial-aware capsules based on cluster centers with spatial encodings, while the previous studies simply build the capsules based on the single representation vector generated by fully-connection or pooling based local feature aggregator. As a result, the previous methods fail to preserve the spatial relationships between local regions, which further limits the representation learning ability of dynamic routing.

In this subsection, in order to efficiently learn the prior logs, we first independently generate the spatial-aware capsules from the feature-spatial embeddings using \emph{rearrange} and \emph{squashing}. Then, we propose to apply routing algorithm between the spatial-aware capsules.

\subsubsection{Rearrange and Squashing}

\begin{figure}[!t]
  \centering
  \includegraphics[width=\linewidth]{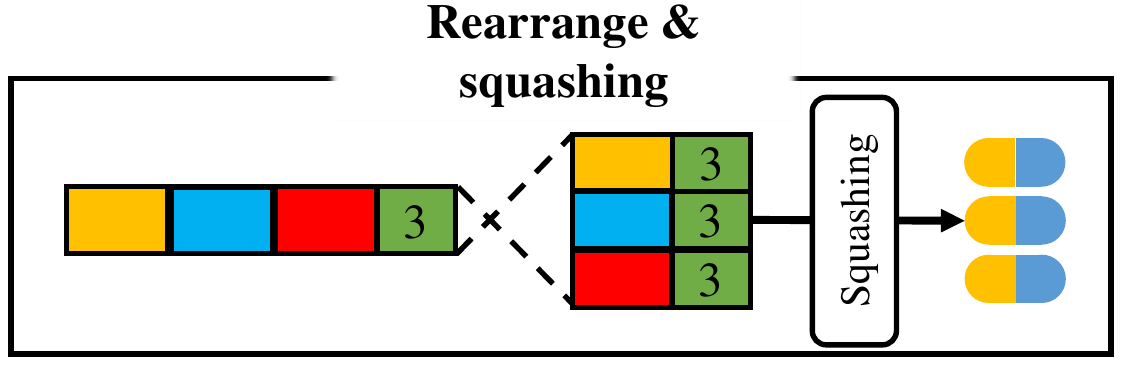}
  \caption{Illustration of rearrange and squashing layer. The green block is the 3D spatial embedding, while the yellow, blue and red block together constitute a feature embedding.}
  \label{fig:spatial-shuffling}
\end{figure}

To build the spatial-aware capsules from the feature-spatial embeddings produced by the geometric feature aggregation module, we deviate from the 2D practice of the original capsule network \cite{sabour2017dynamic}. In the original capsule network, the spatial-aware capsule aggregates its representation vector by collecting the output logits across different channels at the same location on the feature maps. In Point2SpatialCapsule, since we have built the feature-spatial embeddings encoded with the spatial locations, we can consider that each embedding corresponds to a fixed location, which is the learnable cluster center. Therefore, we can directly rearrange the output and use the fully-connected layer with a squashing activation to produce the spatial-aware capsules. The rearrange layer is to split the feature-spatial embeddings $C(\bs{s}_k)$ into several short vectors $\{\bs{u}_i\}$. As shown in Fig. \ref{fig:spatial-shuffling}, the input feature-spatial embedding is split into $K=3$ vectors, each of which is combined with the spatial embedding. Then, we follow a squashing layer, as denote by
\begin{equation}
  \mathrm{squashing}(\bs{u}_i)=\frac{\|\bs{u}_i\|^2}{1+\|\bs{u}_i\|^2}\frac{\bs{u}_i}{\|\bs{u}_i\|},
\end{equation}
where the spatial-aware capsules $\{\bs{u}_i\}$ are generated as the final output of this layer.

\subsubsection{Routing Algorithm}

Given the input spatial-aware capsules, we follow \cite{sabour2017dynamic} to apply dynamic routing algorithm to obtain the digit capsule. Specifically, the digit capsule $\bs{v}_j$ is the output of weighted sum of the prediction vector $\bs{\hat{u}}_{ij}$ followed by the squashing layer, which can be formulated as
\begin{equation}
\label{eq:u_ij}
  \bs{\hat{u}}_{ij} = W_{ij}\bs{u}_{i},
\end{equation}
\begin{equation}
\label{eq:v_j}
  \bs{v}_j = \mathrm{squashing}(\sum_{i}c_{ij}\bs{\hat{u}}_{ij}),
\end{equation}
where $\bs{u}_{i}$ is the $i$th spatial-aware capsule and $W_{ij}$ is a learnable matrix. The \emph{coupling coefficients} \cite{sabour2017dynamic} $c_{ij}$ is determined by the iterative dynamic routing process, denote by
\begin{equation}
\label{eq:c_ij}
  c_{ij} = \frac{e^{b_{ij}}}{\sum_k e^{b_{ik}}}.
\end{equation}

In 2D capsule network, the $\{b_{ij}\}$ are log priors that only depend on the fixed locations and the type of two capsules. In our network, because the disordered input features are clustered as the feature-spatial embeddings by soft-assignment, these features are bounded to the fixed locations (which are the cluster centers) in the feature space. Therefore, the dynamic routing can learn the log priors between these centers and the digit capsules.

Before training, all of the log priors $\{b_{ij}\}$ are initialized to zero. During training, $\{b_{ij}\}$ are learned discriminatively at the same time with other parameters in the network, by adding the scalar product of $\bs{v}_{ij}$ and $\bs{\hat{u}}_{ij}$, i.e.
\begin{equation}
  b_{ij} \leftarrow b_{ij}+ \bs{v}_{ij}\cdot \bs{\hat{u}}_{ij}.
\end{equation}
%

\subsection{Point2SpatialCapsule: Training}
\label{sec:model_description:4}

Following the practice of \cite{sabour2017dynamic}, Point2SpatialCapsule uses the reconstruction loss and the classification loss for supervised point cloud representation learning.

The length of each digit capsule indicates the probability that the characteristic represented by this capsule exists in the input point clouds \cite{sabour2017dynamic}. During training, the margin loss $\mathcal{L}_{cls}$ is adopted for shape classification defined as
\begin{equation}
\begin{split}
   \mathcal{L}_{cls}= & \sum_{j}T_j\max(0,m^{+}-\|\bs{v}_j\|)^2+ \\
     & \sum_{j}T_j\lambda(1-T_j)\max(0,\|\bs{v}_j\|-m^{-})^2,
\end{split}
\end{equation}
where $T_j=1$ if class $j$ is the true label; otherwise, $T_j=0$. $m^{+}$, $m^{-}$ and $\lambda$ are the hyper parameters.

We further reconstruct the input point clouds using four fully-connected layers, with each layer followed by a $relu$ activation and batch normalization except for the last layer. The digit capsule corresponding to the true label is used as the input representation vector to the reconstruction network. The chamfer loss between the original point cloud $\mathbf{X}$ and the reconstructed point cloud $\mathbf{\hat{X}}=\{\hat{\bs{x}}_i\}$ is adopted as the reconstruction loss $\mathcal{L}_{rec}$, as denoted by
\begin{equation}
  \mathcal{L}_{rec}=\frac{1}{|\mathbf{X}|}\sum_{\bs{x}\in \mathbf{X}}\min_{\bs{\hat{x}}\in \mathbf{\hat{X}}}\| \bs{x} - \bs{\hat{x}} \| + \frac{1}{|\mathbf{\hat{X}}|}\sum_{\bs{\hat{x}}\in \mathbf{\hat{X}}}\min_{\bs{x}\in \mathbf{X}}\| \bs{\hat{x}} - \bs{x} \|.
\end{equation}
The total loss for training is the weighted sum of margin loss and the reconstruction loss, as denote by
\begin{equation}
\label{eq:total_loss}
  \mathcal{L}=\mathcal{L}_{cls} + \alpha \mathcal{L}_{rec},
\end{equation}
where $\alpha = 0.0001$ for all the experiments in this paper.

\subsection{Model Adjustments for Part Segmentation}
\label{sec:model_description:5}

\begin{figure}[!t]
  \centering
  \includegraphics[width=\linewidth]{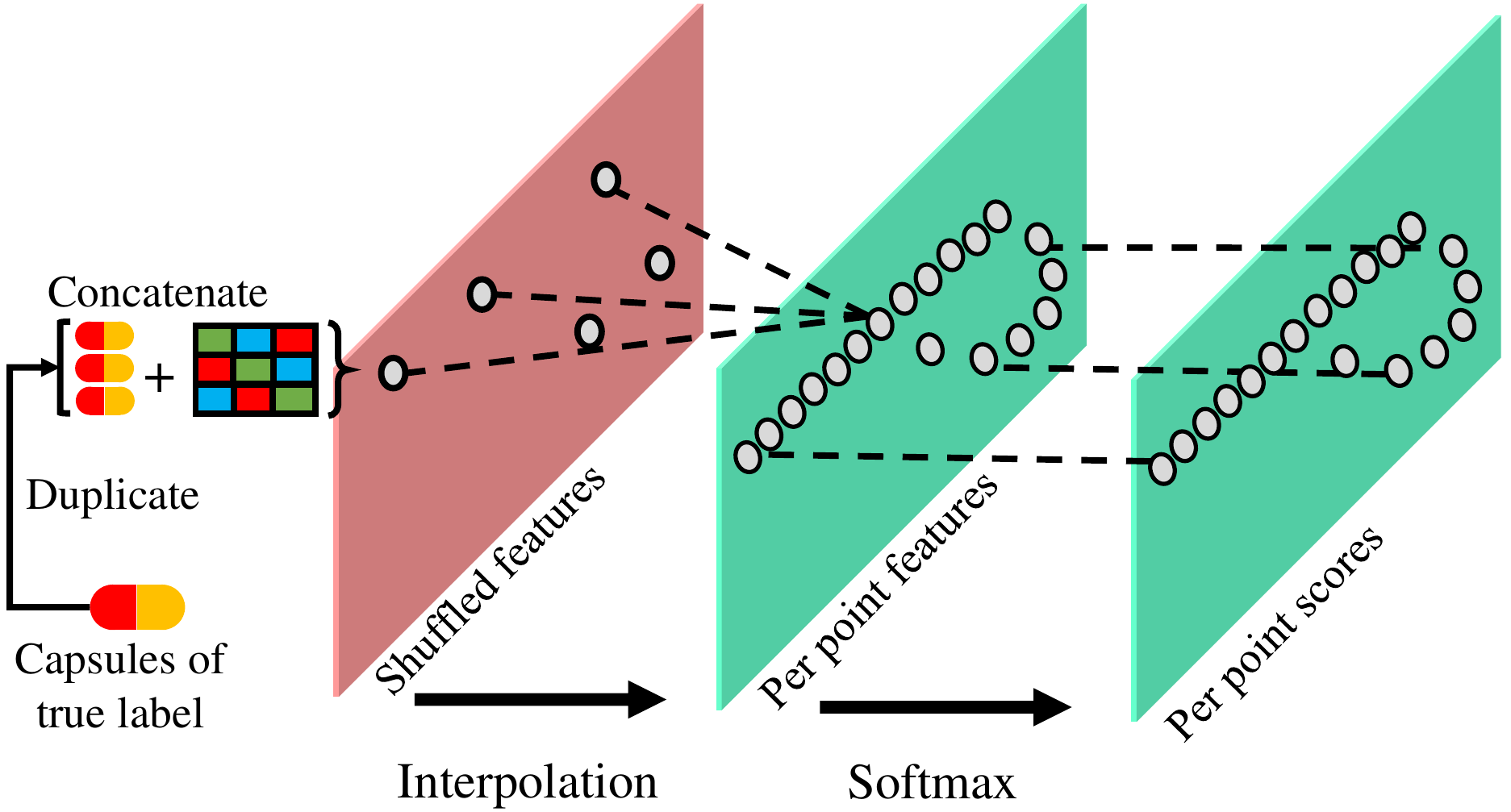}
  \caption{Illustration of the segmentation network in our Point2SpatialCapsule.}
  \label{fig:segmentation}
\end{figure}
The goal of part segmentation is to predict a semantic label for each point in the point cloud.
There are two alternative ways for acquiring the per-point feature for each point from the global feature: duplicating the global feature with $N$ times \cite{qi2017pointnet,wang2018dynamic}, or performing upsampling by interpolation \cite{qi2017pointnet++,li2018so}.
In this paper, we follow the second way to duplicate the vectors in digit capsules belonging to the true label. Then we concatenate the duplicated vectors with the shuffled features. The interpolation layers are used for propagating the features from shape level to point level by upsampling, as shown in Fig. \ref{fig:segmentation}.


\section{Experiments}
\label{sec:experiments}

\subsection{Experimental Setup}
\label{subsec:implementation}

\subsubsection{Datasets}
The 3D shape classification and retrieval experiments are conducted on two subsets of the Princeton ModelNet dataset \cite{wu20153d}, i.e. ModelNet40 and ModelNet10. The ModelNet40 dataset contains 12,311 shapes which belong to 40 categories. We follow the same training and split settings as \cite{li2018so}, which contains 9,843 shapes for training and 2,468 shapes for testing, respectively. The ModelNet10 dataset is a relatively small dataset which contains the 10 common categories of ModelNet40. Following \cite{li2018so}, we split the ModelNet10 into 2,468 training samples and 909 testing samples. Since the original ModelNet provides CAD models represented by vertices and faces, we use the prepared ModelNet10/40 data from \cite{qi2017pointnet++} for fair comparison.
The part segmentation task is conducted on the ShapeNet part dataset \cite{savva2016shrec}, which contains 16,881 models from 16 categories and is split into training, validation and testing following PointNet++.
There are 2048 points sampled for each 3D shape, where each point in a point cloud object belongs to certain one of 50 part classes and each point cloud contains 2 to 5 parts.

\subsubsection{Classification and Retrieval Settings}

Because the length of representation vector in digit capsule indicates the probability that certain characteristic exists in the input point clouds. In the case of Point2SpatialCapsule, the characteristic of digit capsule is the class label. Thus, we choose the digit capsule $\bs{v}_j$ with the biggest length $\|\bs{v}_j\|$ as the predicted label for shape classification.
For the shape retrieval task, we use the Euclidean distances between the length vectors $V=[\|\bs{v}_1\|,\|\bs{v}_2\|,...,\|\bs{v}_m\|]$ of point clouds for similarity measurement. Such similarity measurement is in accordance with the way how capsule stores information.
What's more, a direct comparison between the length vectors requires less computational cost than comparing representation vectors in capsules.

\subsubsection{Implementation Details}

In this paper, we use two multi-scale local feature extraction layer for hierarchically extracting features from point clouds. For the first feature extractor, the input is 1024 points associated with their x, y and z coordinates, from which 512 points is sampled using farthest point sampling. For each sampled point, we select $[8, 16, 32, 64]$ nearest neighbor points of four scales. The MLPs used in the first block have $[32,32,64]$ units for each layer. The second feature extractor samples 256 points out of the 512 points. The number of points for kNN search is the same as the first block. The MLPs for the second block have the units of $[64,64,128]$ for each layer. The parameter $r$ for multi-scale shuffling is 2. The number of the cluster centers is $Q=64$ and the dimension is $C=256$. In the rearrange and squashing layer, we split each embedding into 16 16-dimensional short vectors, which form 1024 16-dimensional spatial-aware capsules in total.

\subsection{3D Shape Classification}

Table \ref{table:cls} compares Point2SpatialCapsule with the existing state-of-the-art methods of point cloud representation learning in terms of shape classification accuracy under ModelNet10 and ModelNet40, respectively. For fair comparison, all the results in Table \ref{table:cls} are obtained under the same input, which handles with raw point sets. Point2Capusule achieves a superior result ($93.4\%$) under ModelNet40, which is higher than the baseline method PointNet++ by $2.7\%$. Specially, Point2Capusule with additional normal vectors achieves the best results ($95.9\%$ and $93.7\%$), compared with the best additional-input method SO-Net \cite{li2018so} ($95.7\%$ and $93.4\%$), under ModelNet10 and ModelNet40, respectively.

We note that both PointNet++ and Point2SpatialCapsule use a multi-scale local feature extraction strategy, where the difference lies in the method used for aggregating local features. The PointNet++ applies max-pooling for aggregating the local features, while Point2SpatialCapsule uses the geometric feature aggregation with spatial relationship aggregation for learning the global representation. Therefore, the improvement in classification accuracy of Point2SpatialCapsule proves the effectiveness of the proposed network for local feature aggregations.

3DCapsule \cite{che2019capsule} is the work most related to our Point2SpatialCapsule in Table \ref{table:cls}. As already discussed in Sec.\ref{sec:related_work}, 3DCapsule simply applies the capsule network on the global features produced by a pooling/full-connected layer, which falls into the scenario of information loss of the spatial locations.

In contrast, our Point2SpatialCapsule applies dynamic routing on the feature-spatial embeddings generated by the geometric feature aggregation module, which can aggregate both the features and their spatial location.
The experimental results in Table \ref{table:cls} shows the implementation of capsule network in our Point2SpatialCapsule is more effective than the implementation of 3DCapsule.

Compared with PointCNN \cite{li2018pointcnn} and DGCNN \cite{wang2018dynamic}, Point2SpatialCapsule still achieves the best results. We note that, PointCNN and DGCNN are also CNN-based neural network, which aims to preserve the spatial locations and spatial relationships of local regions. However, both of them use the max-pooling for aggregating the local region features, which filters out the spatial locations and relationships, especially when aggregating the local features into the global features. As shown Table \ref{table:cls}, the proposed Point2SpatialCapsule yields better performance than the PointCNN and DGCNN, which demonstrates the superior advantages of Point2SpatialCapsule for preserving the spatial locations and relationships.

As seen in Table \ref{table:cls}, our Point2SpatialCapsule outperforms most of the xyz-input methods on point clouds. Specifically, our result is ranked the first place under ModelNet10 ($95.8\%$), and ranked the second place under ModelNet40 ($93.4\%$) which is slightly lower than RS-CNN \cite{liu2019relation} by $0.2\%$. As claimed in \cite{liu2019relation}, RS-CNN performed ``ten voting tests with random scaling and averages the predictions'' during testing. In contrast, we only apply the single model prediction for fair comparison with most of the existing methods \cite{li2018so,che2019capsule,klokov2017escape}. Moreover, when using additional normal vectors as the input, the proposed Point2SpatialCapsule can achieve the best performance among all reported results under ModelNet10 ($95.9\%$) and ModelNet40 ($93.7\%$), respectively. This convincingly verifies the effectiveness of Point2SpatialCapsule.

\begin{table}[!t]
\centering
\caption{The shape classification accuracy (\%) comparison on ModelNet10 and ModelNet40.}
\begin{tabular}{l|c|cc}
\hline
\multirow{2}{*}{Method} &\multicolumn{1}{c|}{\multirow{2}{*}{Input}} &\multicolumn{1}{c}{\multirow{2}{*}{ModelNet10}}&\multicolumn{1}{c}{\multirow{2}{*}{ModelNet40}} \\
    & \multicolumn{1}{c|}{} &\multicolumn{1}{c}{} & \multicolumn{1}{c}{} \\ \hline
 PointNet \cite{qi2017pointnet}  &$1024 \times 3  $  &-  &89.2 \\
 PointNet++ \cite{qi2017pointnet++}	  &$1024 \times 3  $  &-  &90.7 \\
 PointNet++ \cite{qi2017pointnet++}	  &$1024 \times 3$ + norm  &-  &91.9 \\
 SCN \cite{xie2018attentional}	  &$1024 \times 3  $  &-  &90.0 \\
 Kd-Net \cite{klokov2017escape}    &$2^{15} \times 3$   &94.0  &91.8 \\
 KC-Net \cite{shen2018mining}  &$1024 \times 3$        &94.4     &91.0 \\
 PointCNN \cite{li2018pointcnn}     &$1024 \times 3$    &-  &91.7 \\
 DGCNN \cite{wang2018dynamic}     			         &$1024 \times 3$  &- &92.2 \\
 SO-Net \cite{li2018so}  &$2048 \times 3$  &94.1  &90.9 \\
 SO-Net \cite{li2018so}  &$5000 \times 3$ + norm  &95.7  &93.4 \\
 Point2Sequence \cite{liu2019sequence}      &$2048 \times 3$    &95.3  &92.6 \\
 $\Psi$-tree \cite{lei2019octree}      &$2048 \times 3$    &94.6  &92.0 \\
 PATs \cite{yang2019modeling} &$1024 \times 3$  &-    &92.2 \\
 PointWeb \cite{zhao2019pointweb} &$1024 \times 3$  &-    &92.3 \\
 A-CNN \cite{komarichev2019cnn} &$1024 \times 3$  &95.5    &92.6 \\
 RS-CNN \cite{liu2019relation} &$1024 \times 3$  &-    &93.6 \\
 PointConv \cite{wu2019pointconv} &$1024 \times 3$  &-    &92.5 \\
 3DCapusle \cite{che2019capsule}  &$1024 \times 3$    &-    &91.5 \\ \hline
 Point2SpatialCapsule(Ours)      &$1024 \times 3$     &95.8  &93.4 \\
 Point2SpatialCapsule(Ours)      &$1024 \times 3$ + norm     &\textbf{95.9}  &\textbf{93.7} \\ \hline
\end{tabular}
\label{table:cls}
\end{table}

\subsection{3D Shape Retrieval}
\begin{table}[t]
\centering
\caption{The shape retrieval accuracy in terms of mAPs on ModelNet10 and ModelNet40. }
\begin{tabular}{l|c|cc}
\hline
\multirow{2}{*}{Method} &\multicolumn{1}{c|}{\multirow{2}{*}{Input}} &\multicolumn{1}{c}{\multirow{2}{*}{ModelNet10}}&\multicolumn{1}{c}{\multirow{2}{*}{ModelNet40}} \\
    & \multicolumn{1}{c|}{} &\multicolumn{1}{c}{} & \multicolumn{1}{c}{} \\ \hline
 PointNet* \cite{qi2017pointnet} & $1024\times 3$  &67.98 &62.41 \\
 PointNet++* \cite{qi2017pointnet++} &$1024\times 3$ &72.52  &68.97  \\
 3DShapeNets \cite{savva2016shrec} &Multi-View &68.26   &49.23 \\
 DeepPano \cite{shi2015deeppano} &Multi-View &84.18   &76.81 \\
 MVCNN \cite{qi2016volumetric} &Multi-View &-   &83.0 \\
 PANORAMA \cite{sfikas2017panorama} &Multi-View &87.39   &83.45 \\
 GIFT \cite{bai2017gift} &Multi-View &91.12   &81.94 \\
 SequenceViews \cite{han2019seqviews2seqlabels} &Multi-View &89.55  &89.0 \\
 SPNet \cite{yavartanoo2018spnet} &Multi-View &\textbf{94.20}  &85.21 \\
 VIPGAN \cite{han2019view} &Multi-View &90.69  &89.23 \\\hline
 Point2SpatialCapsule(Ours) &$1024 \times 3$ &\textbf{93.43} &\textbf{89.43} \\ \hline
\end{tabular}
\label{table:retrieval}
\end{table}
\begin{figure}[!t]
  \centering
  \includegraphics[width=\linewidth]{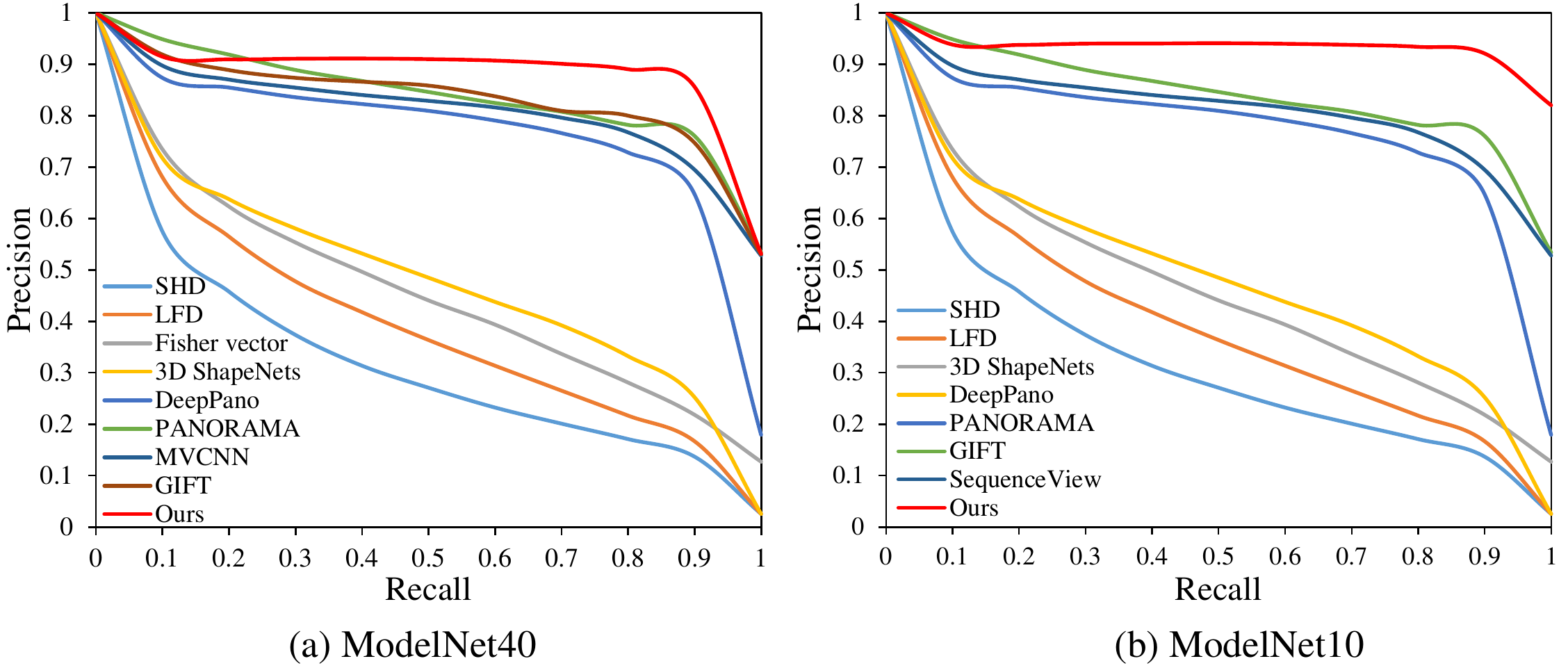}
  \caption{The comparison of precision and recall curves obtained by different methods under (a) ModelNet40 and (b) ModelNet10.}
  \label{fig:PR}
\end{figure}

\begin{table*}[!t]
\caption{The accuracies (\%) of part segmentation on ShapeNet part segmentation dataset.}
\label{table:part_segmentaion}
\begin{tabular}{l|c|cccccccccccccccccccc}
\hline
\multirow{2}{*}{}&
	\multicolumn{1}{c|}{\multirow{2}{*}{mean}} &
	\multicolumn{16}{c}{Intersection over Union (IoU)}\\
	& \multicolumn{1}{c|}{}
    & \multicolumn{1}{c}{air.}
    & \multicolumn{1}{c}{bag}
    & \multicolumn{1}{c}{cap}
    & \multicolumn{1}{c}{car}
    & \multicolumn{1}{c}{cha.}
    & \multicolumn{1}{c}{ear.}
    & \multicolumn{1}{c}{gui.}
    & \multicolumn{1}{c}{kni.}
    & \multicolumn{1}{c}{lam.}
    & \multicolumn{1}{c}{lap.}
    & \multicolumn{1}{c}{mot.}
    & \multicolumn{1}{c}{mug}
    & \multicolumn{1}{c}{pis.}
    & \multicolumn{1}{c}{roc.}
    & \multicolumn{1}{c}{ska.}
    & \multicolumn{1}{c}{tab.}
    \\ \hline
\# SHAPES &   &2690 &76 &55 &898 &3758 &69 &787 &392 &1547 &451 &202 &184 &283 &66 &152 &5271 \\ \hline
PointNet \cite{qi2017pointnet}			&83.7&83.4&78.7&82.5&74.9&89.6&73.0&91.5&85.9&80.8&95.3&65.2&93.0&81.2&57.9&72.8&80.6 \\ \hline
PointNet++ \cite{qi2017pointnet++} &85.1&82.4&79.0&87.7&77.3&90.8&71.8&91.0&85.9&83.7&95.3&{71.6}&94.1&81.3&58.7&{76.4}&82.6 \\ \hline
SCN \cite{xie2018attentional}  &84.6 &83.8 &80.8 &83.5 &{79.3} &90.5 &69.8 &{91.7} &86.5 &82.9 &{96.0 }&69.2 &93.8 &82.5 &{62.9} &74.4 &80.8 \\ \hline
Point2Seq \cite{liu2019sequence}	&{85.2} &82.6 &81.8 &87.5 &77.3 &90.8 &77.1 &91.1 &86.9 &83.9 &{95.7}&70.8&94.6&79.3&58.1&75.2&82.8 \\ \hline
Kd-Net	\cite{klokov2017escape}	&82.3&80.1&74.6&74.3&70.3&88.6&73.5&90.2&87.2&81.0&94.9&57.4&86.7&78.1&51.8&69.9&80.3 \\ \hline
O-CNN \cite{wang2017cnn} &85.2 &84.2 	&86.9 &84.6 &74.1 &90.8 &81.4 &91.3 &87.0 &82.5 &94.9 &59.0 &94.9 &79.7 &55.2 &69.4 &84.2  \\ \hline
KCNet  \cite{shen2018mining} &84.7&82.8&81.5&86.4&77.6&90.3&76.8&91.0&87.2&{84.5}&95.5&69.2&94.4&81.6&60.1&75.2&81.3 \\ \hline

DGCNN \cite{wang2018dynamic} &85.1 &84.2 &83.7 &84.4 &77.1 &90.9 &78.5 &91.5 &87.3 &82.9 &96.0 &67.8 &93.3 &82.6 &59.7 &75.5 &82.0	\\ \hline
SO-Net 	 \cite{li2018so}&84.9&82.8&77.8&{88.0}&77.3&90.6&73.5&90.7&83.9&82.8&94.8&69.1&94.2&80.9&53.1&72.9&{83.0} \\ \hline
RS-CNN 	 \cite{liu2019relation}&86.2 &83.5 &84.8 &{88.8}&79.6 &91.2 &81.1 &91.6 &88.4 &86.0 &96.0 &73.7 &94.1 &83.4 &60.5 &77.7 &{83.6} \\ \hline
PointCNN 	 \cite{li2018pointcnn}&86.1 &84.1 &86.5 &{86.0} &80.8 &90.6 &79.7 &92.3 &88.4 &85.3 &96.1 &77.2 &95.3 &84.21 &64.23 &80.0 &{83.0} \\ \hline
Ours &{85.3}&83.5&83.4&{88.5}&77.6&{90.8}&{79.4}&90.9&86.9&84.3&{95.4}&{71.7}&{95.3}&{82.6}&60.6&75.3&82.5 \\ \hline
\end{tabular}
\end{table*}
\begin{figure*}[!t]
  \centering
  \includegraphics[width=\textwidth]{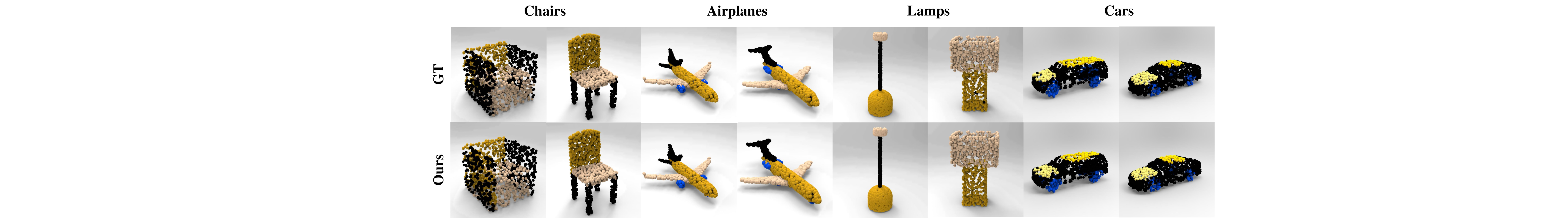}
  \caption{Visualization of part segmentation results. In each shape pair, the first row is the ground truth (GT), and the second row is our predicted result. Parts with the same color are in the same part class.}
  \label{fig:seg_visual}
\end{figure*}

In Table \ref{table:retrieval}, we compare the proposed Point2Capusules with counterpart methods in 3D shape retrieval task, in terms of \emph{mean average precisions} (mAPs). Since most of the methods focusing on 3D shape retrieval are based on multi-views of 3D models, in this subsection, we also quote the experimental results of the multi-view based methods to verify the effectiveness of Point2SpatialCapsule. Note that, the results of PointNet and PointNet++ are obtained by following the same training procedure as described in their original papers, which are denoted by $*$ in this table.

As shown in Table \ref{table:retrieval}, our method has achieved a comparable retrieval accuracy compared with multi-view based methods on both ModelNet10 and ModelNet40. Specifically, Point2SpatialCapsule achieves the best retrieval accuracy $89.43\%$ on ModelNet40, among all reported retrieval results. Point2Capusles achieves the second place result ($93.43\%$) on ModelNet10, which is slight lower than SPNet \cite{yavartanoo2018spnet} by $0.77\%$. However, Point2SpatialCapsule still beats SPNet by $4.22\%$ on ModelNet40 in terms of mAPs, which shows a more balanced performance of Point2SpatialCapsule over different scales of datasets. The comparison of precision-recall (PR) curves under ModelNet40 and ModelNet10 are shown in Fig. \ref{fig:PR}, where the results of Point2SpatialCapsule show the high performance for the 3D shape retrieval task.

The better performance of Point2SpatialCapsule can be dedicated to the following two reasons. First, the Point2SpatialCapsule is able to learn to encode the spatial locations of local features, which can produce a more discriminative representation for point clouds. Second, the digit capsules provide a more interpretable features for representing the point clouds, which is the length vector. Compared with the traditional single vector representations, in which the high-level characteristics are implicitly encoded in the latent feature space, the length of digit capsule explicitly indicates the probability that the characteristics appear in the point clouds.
Therefore, using the distance between length vectors of digit capsules is more effective and interpretable for 3D shape retrieval.

\subsection{3D Shape Part Segmentation}

In Table \ref{table:part_segmentaion}, we also report the performance of Point2SpatialCapsule on the part segmentation task in terms of the Intersection over Union (IoU) \cite{qi2017pointnet}.
As shown in Table \ref{table:part_segmentaion}, our Point2SpatialCapsule achieves the mean instance IoU of $85.3\%$, which outperforms the baseline method PointNet++ on 13 categories out of total 16 categories. Note that, same as PointNet++, Point2SpatialCapsule also employs the multi-scale sampling and grouping strategy for local feature extraction. Therefore, the experimental results prove that Point2Sequence improves the quality of local feature extraction, and leads to the better performance on the segmentation task. Fig. \ref{fig:seg_visual} visualizes some examples of our segmentation results, where our results are highly consistent with the ground truth.

\begin{figure*}[!t]
  \centering
  \includegraphics[width=\textwidth]{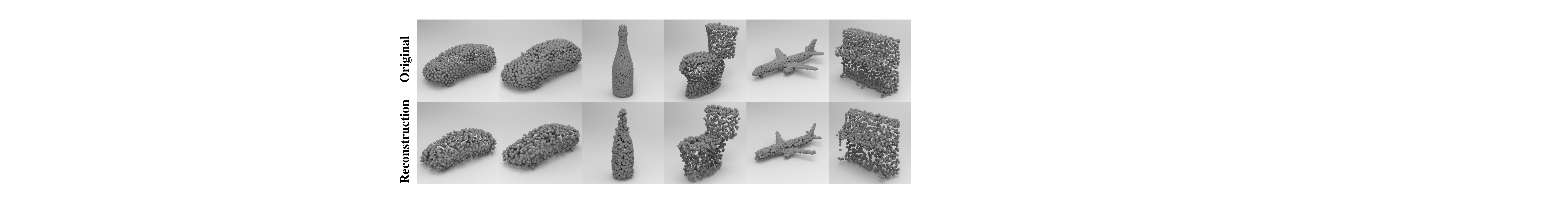}
  \caption{The visualization of reconstruction results on the test set of ModelNet40. The top roll is the input original point cloud, and the bottom roll is the reconstructed point cloud from the Point2SpatialCapsule's reconstruction network. }
  \label{fig:reconstruct}
\end{figure*}
%


Note that, segmentation application needs discriminative features of local regions. Although Point2Capsule is proposed for global shape features by encoding the information of spatial locations in local regions, rather than producing more discriminative features of local regions like RS-CNN \cite{liu2019relation} and PointCNN \cite{li2018pointcnn}, we still achieve comparable results in segmentation results.


\subsection{Ablation Studies}
\label{subsec:ablation}

In this section, we keep the settings of the network the same as described in Sec.\ref{sec:model_description}, except for the specified part for ablation study. We first investigate the influence of each part to our model, and then we analyze three important hyper-parameters in terms of classification accuracy on ModelNet40.
\begin{table}[!htp]
\centering
\caption{The effect of each part of Point2SpatialCapsule on ModelNet40.}
\label{table:shuffle}
\begin{tabular}{C{1.4cm}|cccc}\hline
Model &No-Multi &No-VLAD &No-Caps &Full-Model\\ \hline
Acc (\%)&92.5&91.4&92.1&\textbf{93.44}\\ \hline
\end{tabular}
\end{table}

\subsubsection{The Influence of Each Part to Point2SpatialCapsule}

In order to investigate the effect of each part in Point2SpatialCapsule, we develop and evaluate three different variations of our model as follows. (1) `No-Multi' is the model without multi-scale shuffling, where the output of the multi-scale feature extractor is the direct input to the soft-assignment layer. (2) `No-VLAD' is the model without the geometric feature aggregation, where the output of multi-scale feature extraction layer is directly reshaped as spatial-aware capsules and input to the dynamic routing layer. (3) `No-Caps' is the model without the capsule net, where the output of geometric feature aggregation module is concatenated as a single vector and directly fed into the fully-connected layer for shape classification.

The experimental results are shown in Table \ref{table:shuffle}. From the results we can find that each part of Point2SpatialCapsule contributes to the model performance. We note that, the No-VLAD model achieves the worst performance among the four models, which means that directly applying the capsule network on the point cloud impairs the model's representational ability. The result of No-VLAD model supports our point of view that dynamic routing cannot learn the log priors directly from the disordered point clouds, and verifies the effectiveness of the proposed geometric feature aggregation.
The results of No-Multi proves the importance of applying multi-scale shuffling for smoothing the perceived range between features of different scales. The significant improvement of Full-Model compared to the No-Caps verifies the superior advantage of capsule network for aggregating local features in point cloud recognition.

\subsubsection{The Analysis of Capsules Net}

\begin{table}[!htp]
\centering
\caption{The effect of the iterations of dynamic routing on ModelNet40.}
\label{table:num_capsule}
\begin{tabular}{C{1.4cm}|C{1.4cm}C{1.4cm}C{1.4cm}}\hline
Iterations &{1}&{3}&{5}\\\hline
Acc (\%)&\textbf{93.44} &92.22 &91.98\\ \hline
\end{tabular}
\end{table}

Following the common practice of \cite{sabour2017dynamic}, we investigate the influence of iterations in dynamic routing. As shown in Table \ref{table:num_capsule}, we report the model performance with 1, 3 and 5 iterations of dynamic routing. According to \cite{sabour2017dynamic}, multiple iterations will increase the model's learning ability but may also cause the problem of overfitting. As for Point2SpatialCapsule, we find that dynamic routing with 1 iteration is already enough for learning the point cloud features.

\subsubsection{The Analysis of Geometric Feature Aggregation}

\begin{table}[!htp]
\centering
\caption{The influence of the number of cluster centers in geometric feature aggregation module on ModelNet40.}
\label{table:centers_VLAD}
\begin{tabular}{C{1.4cm}|C{1.2cm}C{1.2cm}C{1.2cm}C{1.2cm}}\hline
Number&{16}&{32}&64&128\\ \hline
Acc (\%) &92.70 &92.93 &\textbf{93.44} &93.21\\ \hline
\end{tabular}
\end{table}

We also analysis the influence of cluster centers in NetVLAD. As shown in Table \ref{table:centers_VLAD}, the model achieves the best result with 64 cluster centers. The explanations are two-fold: (1) the small number of cluster centers could reduce the representational ability of feature embeddings; (2) the slight reduce in performance of the large number of cluster centers is the result of producing similar feature embeddings, which leads to the information redundancy and hinders the model learning more discriminative local features.

\subsubsection{The Analysis of Reconstruction Loss}
\begin{table}[!htp]
\centering
\caption{The influence of reconstruction loss on ModelNet40.}
\label{table:reconstruct}
\begin{tabular}{C{1.4cm}|C{1.2cm}C{1.2cm}C{1.2cm}C{1.2cm}}\hline
$\alpha$&$10^{-3}$&$10^{-4}$&$10^{-5}$&0\\ \hline
Acc (\%) &91.69 &\textbf{93.44} &92.79 &92.46\\ \hline
\end{tabular}
\end{table}

In Table \ref{table:reconstruct}, we discuss the influence of reconstruction loss, where $\alpha$ is the weight factor as specified in Eq. (\ref{eq:total_loss}). From the results, we find that a large $\alpha$ leads to the decreasing of model performance, which in our opinion is the result of a slower learning process cause by the large reconstruction loss weight, especially during the early stage of training.
On the other hand, the experimental results also prove the reconstruction loss useful. Compared with a small weight ($\alpha=10^{-5}$) and the model without reconstruction ($\alpha=0$), the model with $\alpha=10^{-4}$ outperforms them by 0.65\% and 0.98\%, respectively. In Fig. \ref{fig:reconstruct}, we visualize the reconstruction results on the test set of ModelNet40, from which we can find that Point2SpatialCapsule can learn to produce a relatively satisfactory result, despite of the simple reconstruction network employed in the model.

\section{Conclusions}
\label{sec:conclusions}

In this paper, we propose a spatial-aware network, named Point2SpatialCapsule, to jointly aggregate geometric feature and spatial relationships of local regions on point cloud.
The proposed Point2SpatialCapsule has a wide range of potential applications, which can be combined with other local feature extraction methods of multi-scale regions for learning the global shape representation of 3D point clouds.
Compared with the previous feature aggregation methods, Point2SpatialCapsule has the ability to integrate both the geometric features of local regions and the spatial relationships among them. The features of local regions are aggregated by spatial-ware capsules with dynamic routing, which can preserve the spatial relationships between the extracted features.
Experiments show that our network can achieve superior performance on point cloud classification, retrieval and part segmentation tasks under differen datasets.


\bibliographystyle{IEEEtran}
\bibliography{reference}

%

\begin{IEEEbiography}[{\includegraphics[width=1in,height=1.25in,clip,keepaspectratio]{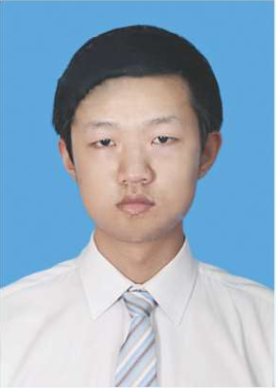}}]{Xin Wen}
received the B.S. degree in engineering management from the Tsinghua University, China, in 2012. He is currently the PhD student with the School of Software, Tsinghua University. His research interests include deep learning, shape analysis and pattern recognition, and NLP.
\end{IEEEbiography}

\begin{IEEEbiography}[{\includegraphics[width=1in,height=1.25in,clip,keepaspectratio]{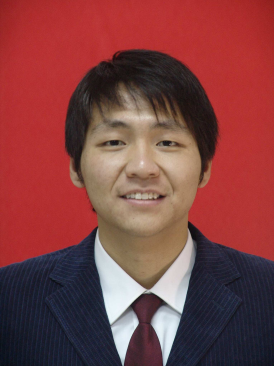}}]{Zhizhong Han}
received the Ph.D. degree from Northwestern Polytechnical University, China, 2017. He is currently a Post-Doctoral Researcher with the Department of Computer Science, University of Maryland at College Park, College Park, USA. He is also a Research Member of the BIM Group, Tsinghua University, China. His research interests include machine learning, pattern recognition, feature learning, and digital geometry processing.
\end{IEEEbiography}

\begin{IEEEbiography}[{\includegraphics[width=1in,height=1.25in,clip,keepaspectratio]{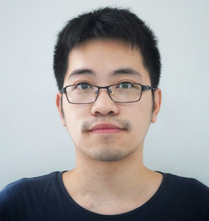}}]{Xinhai Liu}
received the B.S. degree in computer science and technology from the Huazhong University of Science and Technology, China, in 2017. He is currently the PhD student with the School of Software, Tsinghua University. His research interests include deep learning, 3D shape analysis and 3D pattern recognition.
\end{IEEEbiography}

\begin{IEEEbiography}[{\includegraphics[width=1in,height=1.25in,clip,keepaspectratio]{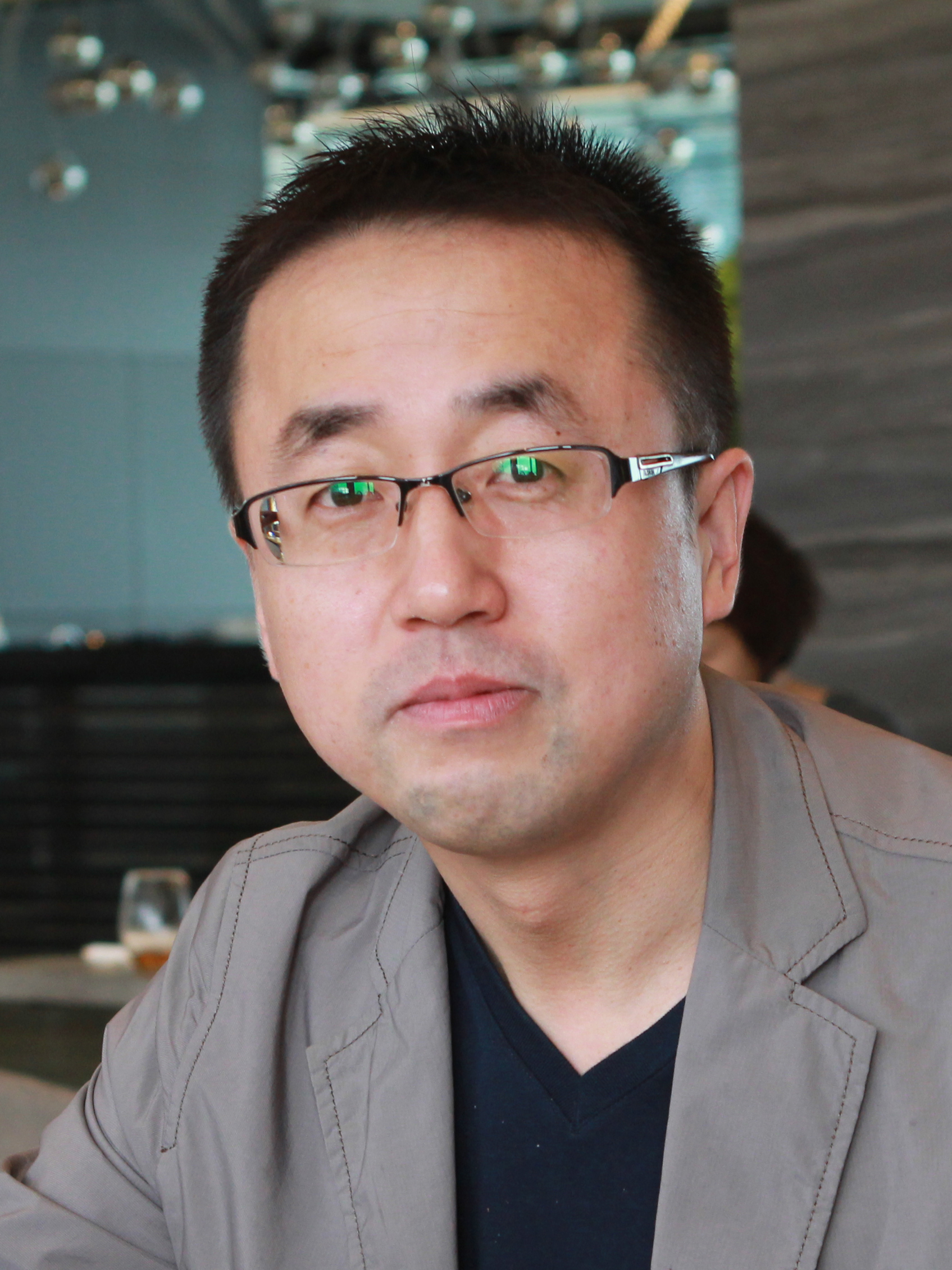}}]{Yu-Shen Liu}
(M'18) received the B.S. degree in mathematics from Jilin University, China, in 2000, and the Ph.D. degree from the Department of Computer Science and Technology, Tsinghua University, Beijing, China, in 2006. From 2006 to 2009, he was a Post-Doctoral Researcher with Purdue University. He is currently an Associate Professor with the School of Software, Tsinghua University. His research interests include shape analysis, pattern recognition, machine learning, and semantic search.
\end{IEEEbiography}






\end{document}